\documentclass[conference]{IEEEtran}
\IEEEoverridecommandlockouts
\usepackage{cite}
\usepackage{amsmath,amssymb,amsfonts}
\usepackage{graphicx}
\usepackage{textcomp}
\usepackage{xcolor}
\usepackage{booktabs}
\usepackage{multirow}
\usepackage{multicol}
\usepackage{subcaption}
\usepackage{algpseudocode}
\usepackage{listings}
\def\BibTeX{{\rm B\kern-.05em{\sc i\kern-.025em b}\kern-.08em
    T\kern-.1667em\lower.7ex\hbox{E}\kern-.125emX}}

\begin{document}

\title{Affinity-Graph-Guided Contractive Learning for Pretext-Free Medical Image Segmentation with Minimal Annotation}

\author{\IEEEauthorblockN{
Zehua Cheng, Di Yuan and Thomas Lukasiewicz}
\IEEEauthorblockA{\textit{
Department of Computer Science
} \\
\textit{University of Oxford}\\
Oxford, United Kingdom\\
zehua.cheng@cs.ox.ac.uk}
}

\maketitle

\begin{abstract}
The combination of semi-supervised learning (SemiSL) and contrastive learning (CL) has been successful in medical image segmentation with limited annotations.
However, these works often rely on pretext tasks that lack the specificity required for pixel-level segmentation, and still face overfitting issues due to insufficient supervision signals resulting from too few annotations.
    Therefore, this paper proposes an affinity-graph-guided semi-supervised contrastive learning framework (Semi-AGCL) by establishing additional affinity-graph-based supervision signals between the student and teacher network, to achieve medical image segmentation with minimal annotations without pretext.
    The framework first designs an average-patch-entropy-driven inter-patch sampling method, which can provide a robust initial feature space without relying on pretext tasks.
    Furthermore, the framework designs an affinity-graph-guided loss function, which can improve the quality of the learned representation and the model's generalization ability by exploiting the inherent structure of the data, thus mitigating overfitting.
    Our experiments indicate that with merely 10\% of the complete annotation set, our model approaches the accuracy of the fully annotated baseline, manifesting a marginal deviation of only 2.52\%. Under the stringent conditions where only 5\% of the annotations are employed, our model exhibits a significant enhancement in performance—surpassing the second-best baseline by 23.09\% on the dice metric and achieving an improvement of 26.57\% on the notably arduous CRAG and ACDC datasets.
\end{abstract}

\section{Introduction}

The precise delineation of medical imagery furnishes pivotal and discerning data for medical practitioners for suitable diagnostic evaluations, monitoring disease evolution, and formulating effective treatment strategies.
Supervised methods based on deep learning have achieved a remarkable performance in medical image segmentation~\cite{Chen2016DeepLabSI}.
However, these methods largely benefit from extensive annotation datasets~\cite{wang2022uncertainty}, and acquiring pixel-level annotations on a broad scale frequently demands a significant time investment and specialized knowledge, and entails substantial expenses.
To alleviate the dependence on a large amount of annotated data, semi-supervised learning (SemiSL) and contrastive learning (CL) complement each other and are widely used in medical image segmentation.
In detail, the pseudo-labels generated by SemiSL enhance the discriminative ability of CL by providing supplementary guidance for the metric learning method~\cite{arazo2020pseudo}, while the crucial class discriminative feature learning of CL enhances the multi-class segmentation efficacy of SemiSL, allowing SemiSL to produce more ideal pseudo-labels~\cite{He2019MomentumCF,chaitanya2020contrastive}.

However, these methods have two obvious shortcomings: (i) \textbf{Relying on pretext tasks leads to a poor generalization ability}.
First, these methods suffer from sampling biases and exacerbated class collision~\cite{Chuang2020DebiasedCL} that undermine the model's performance. 
Then, these methods do not account for substantial domain differences, resulting in a poor performance across datasets for models trained well in pretext tasks~\cite{Gu2022ContrastiveSL}, which is particularly prevalent in medical images.
(ii)~\textbf{The lack of supervision signal leads to an overfitting problem}.
Most of these methods use regression, pixel-wise cross entropy or mean square error loss terms, and their variants to evaluate and generate ideal pseudo labels to assist the model in generating relatively accurate segmentation results.
However, the loss functions have notable limitations, i.e., they cannot enforce intra-class compactness and inter-class separability~\cite{Wu2022ExploringSA,li2023modeling}, thus limiting their full learning potential. 
Besides, there is a domain shift problem, i.e., these methods employ self-integration strategies and are designed for a singular dataset~\cite{wu2022mutual}, which brings challenges to generalization across different domains.


The essence of solving the above problems lies in further utilizing the feature consistency in the manifold space with respect to the regional feature interconnections.
Essentially, the effectiveness of SemiSL is based on the manifold assumption and the cluster assumption~\cite{Yang2021ASO}, which conceptualizes data points as components of low-dimensional manifolds embedded within a larger, high-dimensional space. Data points situated in the same feature space possess identical labels.
However, when dealing with a limited amount of labeled data, the demarcation of cluster boundaries becomes ambiguous, which impedes the accurate delineation of the manifold's shape, leading to difficulties in correctly assigning labels to unlabeled data and thus hurting the quality of the learned feature representation.
To mitigate these issues, the construction of a semantic graph presents a viable solution. By representing data points as nodes and their interconnections as edges based on feature similarities, a semantic graph encapsulates the intricate relationships within the data with explicitly representation.
This structure not only enhances the understanding of manifold geometry but also provides a more nuanced view of cluster boundaries, even in scenarios with minimal labeled data.
The integration of a semantic graph into SemiSL can be instrumental in exploiting the manifold's inherent structure, thereby facilitating more accurate label propagation in sparsely annotated environments.


Therefore, in this work, we propose an affinity-graph-guided semi-supervised patch-based CL framework that avoids the displayed pretext task.
Specifically, the framework first uses an average-patch-entropy-driven new inter-patch semantic disparity mapping to select the positive and negative patches, and to improve sampling within our CL method, thus providing a powerful initial feature space by avoiding class conflicts.
Then, the framework introduces affinity-graph supervision as an external constraint on the pseudo-label generated by the student and teacher networks, to enrich the supervision signal and enhance the discriminative ability required for accurate segmentation by exploiting the inherent structure of the data.
Finally, the framework proposes a new hard negative sampling method, making hard negative samples similar to positive samples but with different labels, and designs a loss function based on the affinity graph between positive and negative samples, to combine the advantages of SemiSL and CL, which can improve the quality of the learning representation and the model's generalization ability.

The main contributions of our proposed method are listed below:
\begin{itemize}
    \item To alleviate the problem of the reliance on pretext and the overfitting problem caused by the lack of supervision signals, we propose an affinity-graph-guided semi-supervised contrastive learning framework (Semi-AGCL) to achieve high-precision medical image segmentation with extremely few annotations.
    \item We use an affinity graph as an external constraint on the generated pseudo labels with our affinity mass loss to minify class-discriminative features without any explicit training on pretext tasks, thereby demonstrating generalizability across multiple domains.
    Furthermore, we utilize a patch-based CL framework, wherein the selection of positive and negative patches is steered by an entropy-based metric, informed by the pseudo-labels garnered in the SemiSL setting. This approach averts class collision, i.e., the forceful and unguided contrasting of semantically akin instances within the CL framework.
    \item Following evaluation across three datasets from diverse domains, our method has demonstrated effectiveness, showcasing its generalizability and robustness even with a limited number of annotated samples.
\end{itemize}

\begin{figure*}[!t]
\centering
\includegraphics[scale=.48]{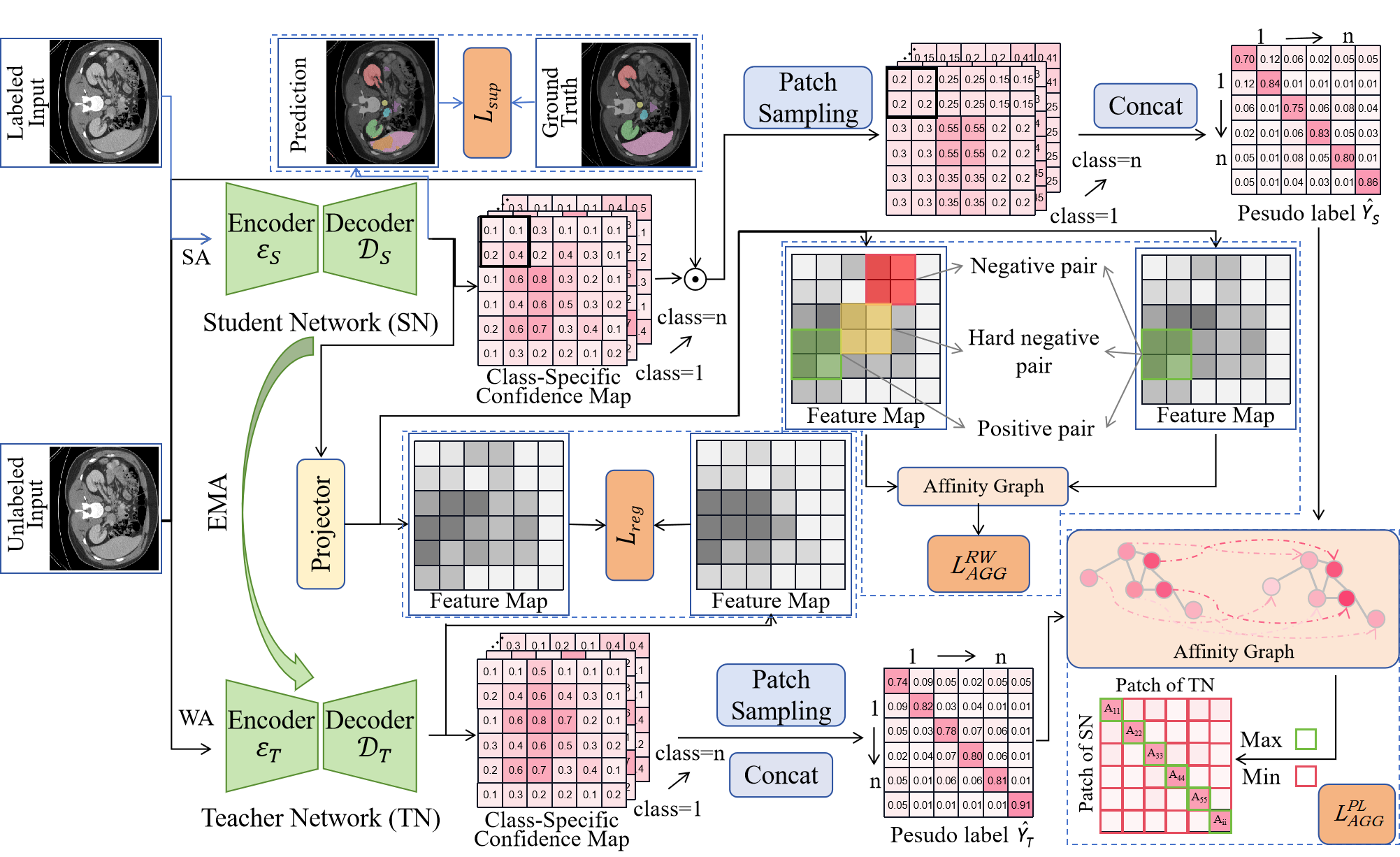}
\caption{The  proposed framework. For labeled data, we directly use the supervised loss $\mathcal{L}_{sup}$ to update the student network. For unlabeled data, we first slice the image into patches, then bridge an affinity graph loss $\mathcal{L}_{AGG}^{PL}$ between pseudo labels of student and teacher networks, and also design a new loss $\mathcal{L}_{AGG}^{RW}$ using the reweighting hard negative sample based on the edge of affinity graph. In the affinity-graph-based losses, we use low $A_{ii}$ to construct a negative hard sample and try to pull positive pairs closer (increase $A_{ii}$) and push negative pairs away. Besides, the blue arrows use labeled data, and the rest (black arrows) are unlabeled data; we use a mixture of labeled and unlabeled data, so it is a semisupervised task rather than a self-supervised task.
SA: Strong Augmentation, WA: Weak Augmentation.}
\label{fig:general}
\end{figure*}

\section{Related Works}

\textbf{Semi-supervised learning (SemiSL)}
harnesses valuable representations from a vast array of unlabeled samples, concurrently with supervised learning on a handful of labeled samples, which encompasses pseudo labeling~\cite{Seibold2021ReferenceguidedPG}, consistency regularization~\cite{Basak2022AddressingCI,Jin2022SemisupervisedHI}, and entropy minimization~\cite{Hang2020LocalAG,Vesal2021AdaptEU}.
The most common one is the pseudo-labeling method based on the Mean Teacher framework.
UA-MT~\cite{Yu2019UncertaintyawareSM} exploits the uncertainty information of the teacher model to guide the student model to learn from meaningful and reliable targets.
Double-UA~\cite{Wang2020DoubleUncertaintyWM} uses a double-uncertainty weighted method to make the teaching-learning process accurate and reliable.
SASSNet~\cite{Li2020ShapeawareS3} incorporates a flexible geometric representation to enforce a global shape constraint and handle objects with varying poses or shapes.
DTC~\cite{Luo2020SemisupervisedMI} jointly predicts a pixel-wise segmentation map and a geometry-aware level set representation of the target.
URPC~\cite{Luo2020EfficientSG} designs an uncertainty rectifying module to enable the framework to learn from meaningful and reliable consensual regions at different scales of pyramid predictions.
MC-Net~\cite{Wu2021SemisupervisedLA} designs a cycled pseudo label scheme between the prediction discrepancies of two decoders to encourage mutual consistency.
SS-Net~\cite{Wu2022ExploringSA} explores pixel-level smoothness and inter-class separation at the same time.

\textbf{Contrastive learning (CL)}
can increase the mutual information of similar samples by maximizing the similarity of positive samples and minimizing the similarity of negative samples, so it has been widely used in computer vision~\cite {chen2020simple,Chen2020BigSM}.
However, it also faces some challenges when applied to medical image segmentation.
First, sampling bias and exacerbated class collision will be led due to the uninformed and unguided selection of negative samples during contrastive learning~\cite{Chuang2020DebiasedCL}, which degrades the discriminative ability of the learned representations, thus hurting the segmentation performance~\cite{Tang2023LACLLC}.
Then, a common practice in CL is to transfer well-trained models in the pretext task of large-scale natural image datasets to downstream tasks in the medical image field~\cite{Zhang2020ContrastiveLO}. However, significant domain shifts across such heterogeneous datasets often have a negative impact on the final task performance~\cite{Zhao2022MTUDATU}.
Lastly, designing an appropriate pretext task itself can be an arduous exercise, and the choice of the pretext task may not generalize well across datasets~\cite{Misra2019SelfSupervisedLO}.

Therefore, how to overcome the limitations of CL in semi-supervised learning with limited labeled data and effectively utilize its representation capabilities for medical image segmentation remains an open challenge. 
Our work is the first attempt to alleviate this gap by proposing an affinity-graph-guided semi-supervised patch-based CL framework that synergistically combines CL and SemiSL through joint optimization.
Unlike existing techniques, our framework does not require any additional pre-training and can be trained end-to-end for medical image segmentation.

\section{Methodology}
Given a labeled image set alongside its respective label set $D_L$ and an unlabeled image set $D_U$, comprising $N_L$ and $N_U$ images (where $N_L<<N_U$), respectively, we propose a patch-wise contrastive learning strategy with a teacher-student model to target the assimilate information from both $D_L$ and $D_U$ directed by pseudo-labels.
In our framework, we first delineate the patch generation process, steered by the effective employment of (true or pseudo) labels; we then devise a new contrastive loss function incorporating an affinity graph between pseudo labels of student and teacher networks; subsequently, we construct an affinity graph between the positive and negative samples to guide the student network to learn from diverse distributions.


\subsection{Patch-wise Class-centric Sampling\label{sec:patch_sampling}}
Let $X_i \in \mathbb{R}^{M}$ denote the $i^{th}$ image in a mini-batch, containing $M$ pixels. The value of the $m^{th}$ pixel in image $X_i$ is denoted by $X_i(m)$. 
The key idea behind our patch-wise class-centric sampling is to select positive and negative patches for contrastive learning in an informed manner using the pseudo-labels. This prevents forceful contrasting of semantically similar patches, i.e., class collision. 
To achieve this, we first generate a class-specific confidence map $C_i^k$ for each class $k \in \{1,2,...,K\}$, where $K (\geq 1)$ indicates the total number of classes.
This confidence map reflects the likelihood of each pixel belonging to class $k$. By performing an element-wise product between $C_i^k$ and the image $X_i$, which accentuates the regions relevant to class $k$, i.e., $X'^k_i = X_i \odot C^k_i$. Although the confidence map $C_i^k$ is much more informative comparing to the segmentation mask. $X'^k_i$ retains the image intensity/texture information along with masking information from the groundtruth and provides a richer representation for entropy calculation.

To sample informative patches, we compute an average patch entropy $Ent^k_{i,j}$ for each patch $P^k_{i,j}$ based on the pixel intensity values in the attended image $X'^k_i$. This entropy reflects three key types of information: confidence of belonging to class $k$, uncertainty regarding other classes, and intensity appearance from the original image $X_i$. A high entropy value indicates the patch likely contains the class $k$ object but also has some confusion with other classes. The entropy thereby provides a richer metric for sampling, instead of simple random selection. 
This guided sampling focuses the learning on informative patches. 
Therefore, the average patch entropy allows robust, semantically meaningful sampling of positives and negatives to serve as a highly informative supervision signal to the self-supervised learning model.
 $Ent^k_{i,j}$ is formulated as follows:
\begin{equation}\label{eq:select_pos}
\begin{split}
    Ent^k_{i,j} = -\frac{1}{|P^k_{i,j}|} &\sum_{m \in P^k_{i,j}} X'^k_i(m)\log(X'^k_i(m)) \\
    &+ (1-X'^k_i(m))\log(1-X'^k_i(m)),
\end{split}
\end{equation}

\noindent
where $X'^k_i(m)$ is the intensity value of pixel $m$ in patch $P^k_{i,j}$. 
For an anchor patch of class $k$, patches with Top-$n$ $Ent^k_i$ values are \textbf{positives}, and the rest are \textbf{negatives}. This entropy-based sampling allows sampling positives that have high confidence for class $k$ while also sampling challenging negatives from other classes. The patch appearance information also helps avoid ambiguity.


\subsection{Affinity-Graph-Guided Contrastive Loss between Pseudo Labels\label{sec:agcl_pseudo_labels}}
Unlike traditional semi-supervised learning frameworks, a patch-wise approach necessitates the incorporation of regional information to maximize the utility of the data. Consequently, it is our contention that not all patches should be regarded as equally significant. Therefore, we introduce an affinity graph to regularize patch importance by constructing fine-grained alignment in the outputs of student network ($\hat{\mathbf{Y}}_S$) and teacher network ($\hat{\mathbf{Y}}_T$).
By directly encoding prediction vector similarities as edge weights between graph nodes, the discrete topology inherently captures the continuous semantic affinities that we intend to align.
Concurrently, the graph Laplacian regularization enforces smoothness priors, forefending collapse into trivial solutions.
Maximizing the resultant diagonal trace impels convergence of the patch-wise pseudo-labels. 

Specifically, we construct a patch-wise affinity graph $\mathbf{A} \in \mathbb{R}^{N\times N}$ between the pseudo-labels from the teacher network $\hat{\mathbf{y}}_t$ and student network $\hat{\mathbf{y}}_s$, where $N$ is the number of patches. The edge weight $A_{ij}$ is defined using a Gaussian kernel based on the $L_2$ distance between pseudo-label vectors $\hat{\mathbf{y}}_t^i$ and $\hat{\mathbf{y}}_s^j$:
\begin{equation}\label{eq:pl_ag}
    A_{ij} = \exp\left(-\frac{|\hat{\mathbf{y}}_t^i - \hat{\mathbf{y}}_s^j|_2^2}{2\sigma^2}\right),
\end{equation}

\lstset{language=Python, 
        basicstyle=\ttfamily\small, 
        keywordstyle=\color{blue},
        commentstyle=\color{teal},
        stringstyle=\color{red},
        showstringspaces=false}

\begin{figure*}
\begin{lstlisting}[caption={The PyTorch implementation of Equation~\ref{equ:PL}. This implementation is based on the PyTorch~\cite{Paszke2019PyTorchAI} and einops~\cite{Rogozhnikov2022EinopsCA} libraries.}\label{algo:pytorch_impl}]
import einops
import torch

def pl_affinity_graphs(student_preds, teacher_preds, gamma):
    # student_preds  [N, H, W, C]
    # teacher_preds  [N, H, W, C]
    s_preds = einops.rearrange(student_preds, 'N H W C -> N (H W) C') # [N, H*W, C] 
    t_preds = einops.rearrange(teacher_preds, 'N H W C -> N (H W) C') # [N, H*W, C]
    dist_mat = torch.cdist(s_preds, t_preds) # [N, H*W, H*W]
    return  torch.exp(-dist_mat / (2 * (gamma**2)))
\end{lstlisting}
\end{figure*}


\noindent
$\sigma$ is an adaptive bandwidth parameter.
We choose the Gaussian kernel, because it has strictly localized support, smooth variation, and efficient computability, which are theoretically well-founded for representing granular semantic relationships among samples.
First, locality is imparted through the exponential term that precipitously decays affinity weight as distance in the embedding hyperspace grows. This realizes the expectation that closer samples exhibit greater semantic similarity and relatedness. 
It also guarantees gradual weight transitions with distance alterations, regulated by $\sigma$, thereby preventing abrupt changes. The modulating impact of $\sigma$ further provides control over the rate of falloff and spatial scope of similar neighborhoods. 
Additionally, the Gaussian kernel satisfies constraints of radial symmetry and positive semi-definiteness suitable for modeling sample-wise relationships rather than discrete differences. Efficient computability facilitates constructing weighted graphs over large corpora encompassing tens of thousands of nodes. 

The affinity graph construct provides a prudent approach here, as directly encoding prediction vector similarities as edge weights inherently captures the desired semantic affinities to align. Concurrently, the discrete topology is regulated through smoothness priors.
In other words, we harness the diagonal entries $A_{ii}$, measuring self-similarities between teacher and student pseudo-labels on patch $i$. By maximizing the trace $\mathbb{E}[\text{tr}(\mathbf{A})]$, we promote convergence of the patch-wise student and teacher predictions.
Simultaneously, we minimize the nuclear norm $\lVert\mathbf{A}\rVert _*$ via convex relaxation. 

The nuclear norm serves as a convex lower bound on the intractable matrix rank function. We first perform this convex relaxation through singular value thresholding~\cite{Cai2008ASV}. This decomposes $\mathbf{A}$ into $\mathbf{A}= \mathbf{U} \Sigma \mathbf{V}^{T}$ via singular value decomposition, where $\Sigma$ contains the singular values $\sigma_i$. We then soft-threshold these singular values by an amount proportional to the subgradient of the nuclear norm, iteratively zeroing out unimportant dimensions. Thereby, minimizing $||\mathbf{A}||{*}$ serves as an efficient, tractable proxy for minimizing rank.

In summary, by excavating this salient low-rank alignment pattern between student and teacher outputs amidst noisy inconsequential variations, our approach safeguards the model from overfitting during contrastive learning. In effect, the convex relaxation extracts the most essential signals while filtering out extraneous dimensions.
Explicitly, we get the affinity graph guided contrastive loss $\mathcal{L}_{\text{PL}}$ between the pseudo labels of $\hat{Y}_S$ and $\hat{Y}_T$ with its affinity graphs $\mathbf{A}$ as:

\begin{equation}
    \mathcal{L}_{\text{AGG}} ^{PL}=\sum^N_{i=1} \exp\left(-\frac{|\hat{\mathbf{y}}_t^i - \hat{\mathbf{y}}_s^i|_2^2}{2\sigma^2}\right)+ \gamma ||\mathbf{A}||*,
    \label{equ:PL}
\end{equation}
where $\gamma$ balances the relative importance. By directly encoding alignment similarities and extracting low-rank structure, this loss function applied alongside standard supervised objectives aligns student and teacher representations robustly even with few labels. The affinity graph topology provides an interpretable, flexible mechanism for semi-supervised contrastive learning.
We note that the quantities needed for the affinity graph in Equation~\ref{eq:pl_ag} implementation are easily computed in parallel across examples within a batch using deep learning. We show a code snippet in Listing~\ref{algo:pytorch_impl}.

\subsection{Affinity-Graph-Guided Hard-Negative Reweighting}
\label{sec:reweighting}
\cite{Robinson2020ContrastiveLW,Jiang2022SupervisedCL} argue that the performance of contrastive learning could be improved by the incorporation of hard negative samples (i.e., samples $y_i$ that are difficult to distinguish from an anchor $x_i$).
In this context, rather than considering two arbitrary data points as negative pairs, these methods construct a negative pair from two random data points that are not too far from each other. The affinity graph constructed by Equation~\ref{eq:pl_ag} also can meassure the hard negative samples under $L_2$ distance between $\hat{\mathbf{y}}^i_t$ and $\hat{\mathbf{y}}^i_s$. Therefore, utilizing the hard negative samples selected from Equation~\ref{eq:pl_ag} could further improve the performance of the current framework.

In contrastive learning, we get the query $\mathbf{q}$ and the corresponding key $\mathbf{k}$ embeddings from the positive pair. 
We construct the query-key paris $(\mathbf{q},\mathbf{k})$ with the encoder-projection routine for the student networks, where we get $\mathbf{q}=\mathcal{H}(\mathcal{E}_S(P_i^k))$, $\mathcal{E}_S$ is the encoder of student networks, and $\mathcal{H}$ is the projection head.
The key vector set $\mathcal{K}$ is formulated by amalgamating both positive and negative keys, represented as $\mathcal{K}= \mathcal{K}^+ \cup \mathcal{K}^-$, $\mathcal{K}^+$ consisting of positive keys $\mathbf{k}^+_i$ with the same distribution as $\mathbf{q}_i$, $\mathcal{K}^-$ consisting of negative samples $\mathbf{k}_i^-$. 
A widely recognized and effective loss function utilized in contrastive learning is delineated as follows:
\begin{equation}
    \label{eq:cl_mem_bank}
    \mathcal{L}_{\mathbf{q},\mathbf{k}^+,Q}=-\log \frac{\exp(\frac{\mathbf{q}^T \cdot \mathbf{k}^+}{\tau})}{\exp(\frac{\mathbf{q}^T \cdot \mathbf{k}^+}{\tau}) + \sum\limits_{\mathbf{n} \in Q} \exp(\frac{\mathbf{q}^T \cdot \mathbf{n}}{\tau})},
\end{equation}

\noindent
where $\tau$ is a temperature parameter. The positive pair $(\mathbf{q},\mathbf{k}^+)$ is contrasted with every feature $n$ in the bank of negatives $Q$~\cite{He2019MomentumCF} with a fixed size $K$.

The log-likelihood function of Equation~\ref{eq:cl_mem_bank} is delineated based on the probability distribution, which emerges from the application of the softmax function to each $\mathbf{q}$. Let $p_{\mathbf{z}_i}$ as the correspondence probability between the query and feature $\mathbf{z}_i\in Z=Q\cup {\mathbf{k}^+}$. In this case, we get $p_{\mathbf{z}_i}=\frac{\exp{\mathbf{q}^T\cdot \mathbf{z}_i/\tau}}{\sum_{j=Z}\exp{\mathbf{q}^T\cdot \mathbf{z}_j/\tau}}$.
The derivative of the loss function relative to the $\mathbf{q}$ is presented as:
\begin{equation}
    \frac{\partial{\mathcal{L}_{\mathbf{q},\mathbf{k}^+,Q}}}{\partial{\mathbf{q}}}= -\frac{1}{\tau}((1-p_k)\cdot \mathbf{k}^+-\sum_{\mathbf{n}\in Q}p_n\cdot \mathbf{n}),
    \label{eq:derivative}
\end{equation}

\noindent
where $p_k$ and $p_n$ represent the probabilities of matching the key and negative feature, respectively. This formulation encapsulates the likelihoods of the feature vector $\mathbf{z}_i$ being aligned with either the key or the negative feature in the given context. It is trivial that the impact of both positive and negative logits on the loss function mirrors that observed in a $(K +1)$-way cross-entropy loss. In this scenario, the logit corresponding to the key is indicative of the latent class of the query. Additionally, all gradients in this framework are uniformly scaled by a factor of $\frac{1}{\tau}$. Therefore, sampling effective hard-negative samples from the memory bank $Q$ has become the most effective way to improve the learning Equation~\ref{eq:cl_mem_bank}.

We sample the hard negative samples based on two principles. First, the sampled hard negative instances must possess labels that are distinct from those of the anchor instances. This criterion ensures the maintenance of a fundamental dissimilarity at the label level. In our framework, for the same patch $P^k_i$ with different views of augmentation, a hard sample/patch would expect to have the lowest similarity  score between the two views.
Then, the most advantageous hard negative samples are those which, according to the current state of the embedding, appear to be similar to the anchor. This perceived similarity, albeit misleading, renders these samples particularly challenging and, therefore, intrinsically valuable for the training process. Such samples, by virtue of their difficulty, provide a robust mechanism for enhancing the discriminative capability of the learned representations.

Based on the abovementioned principles, we set a dynamic threshold $\theta\in (0,1)$ to identify hard negative samples with low edge weight $A_{ii}$ based on the diagonal element of $\mathbf{A}$ constructed by Equation~\ref{eq:pl_ag}. 
Motivated by~\cite{kalantidis2020hard} which proved the effectiveness of a mixture of the query and hard negative sample with a joint projection function.
For the negative features $\mathbf{n}\in Q$ from a memory bank $Q$ of size K, we get:
\begin{equation}
    \mathbf{h}_k=\frac{A_{ii}\mathbf{n}_i+(1-A_{ii})\mathbf{n}_j}{||A_{ii}\mathbf{n}_i+(1-A_{ii})\mathbf{n}_j||_2},
\end{equation}

\noindent
where $||\cdot||_2$ is the $l_2$-norm. $\mathbf{n}_i,\mathbf{n}_j\in Q$ are randomly chosen negative features from the set $Q$ of the closest $N$ negatives. $H$ is the hard negative samples set where $\mathbf{h}_k\in H$.
The $A_{ii}$ represents the diagonal element of the affinity graph, and the suffix $j$ represents other negative features in $Q$.
We now get our affinity-graph-guided hard-negative reweighting on Equation~\ref{eq:cl_mem_bank} as:

\begin{equation}
    \mathcal{L}_{\text{AGG}} ^{RW}=-\log\frac{\exp(\frac{\mathbf{q}^T\cdot \mathbf{k}^+}{\tau})}{\exp(\frac{\mathbf{q}^T\cdot \mathbf{k}^+}{\tau})+\sum\limits_{\mathbf{\mathbf{h_k}}\in H}\exp(\frac{\mathbf{q}^T\cdot \mathbf{h_k}}{\tau})}.
\end{equation}
Since Equation~\ref{eq:cl_mem_bank} calculates the $L2$ distance between the positive sample and negative samples, so only the diagonal element of the affinity graph $A_{ii}$ is used here.

\noindent
Therefore, the overall loss function is as follows:
\begin{equation}
    \mathcal{L}_{all} = \mathcal{L}_{sup} + \mathcal{L}_{reg} + \mathcal{L}_{AGG}^{PL} + \mathcal{L}_{AGG}^{RW},
\end{equation}

\noindent
where $\mathcal{L}_{reg}$ is the cross entropy loss between the outputs of student and teacher networks.

\section{Experiments}
\subsection{Dataset and Metrics}
To assess the performance of our proposed method, we conducted experiments using three different datasets, each representing a distinct input modality. We processed 3D data from the LA dataset, 2D images from the ACDC dataset, and whole slide images from the CRAG dataset. This comparative analysis allowed us to evaluate how our method performs across varying input types. To verify the effectiveness of our proposed method, we deliberately chose a scenario with a small amount of data and limited supervisory signals. Robust performance demonstrated with constrained datasets and limited supervisory signals suggests promising scalability to larger datasets and enhanced supervisory signals. 

The \textbf{LA dataset} is an Atrial Segmentation Challenge dataset~\cite{xiong2021global} including $100$ 3D gadolinium-enhanced magnetic resonance image scans with labels. 
%
The \textbf{ACDC dataset}~\cite{bernard2018deep} is a public segmentation dataset with four classes, i.e., background, right ventricle, left ventricle, and myocardium, containing $100$ patients’ scans.
%
The \textbf{CRAG dataset} is a Colorectal Adenocarcinoma Gland dataset~\cite{Graham2018MILDNetMI} containning $213$ H\&E WS histopathological images taken with an OmnyxVL120 scanner. It has images with $20\times$ objective magnification with a resolution of $0.55 \mu m/pixel$ with each tile possessing full instance-level annotation.

We employ four metrics to evaluate the framework performance on all datasets, namely, Dice Similarity Score (DSC), Jaccard Index (Jaccard), Hausdorff Distance 95 (HD95), and Average Symmetric Distance (ASD)~\cite{chaitanya2020contrastive}.
Given the outputs and the ground truths, DSC and Jaccard mainly evaluate the overlap value between them, HD95 measures the closest point distance between them, and ASD computes the average distance between their boundaries.

We follow the official setup for the training and testing split of all the datasets: both LA and ACDC are 80\% and 20\% for training and validation; and CARG is 80\%, 10\%, and 10\% for training, validation, and testing.

\subsection{Implementation Details}
We conduct all experiments on a DGX A100 server with fixed random seeds. 
The model's convergence is achieved through the utilization of an ADAM optimizer, with the specifications of a batch size set at $16$ and a learning rate designated at $1e-4$. The parameters $\tau$ and $\lambda$ in Equation~\ref{eq:derivative} are assigned values of $0.2$ and $4$, respectively, as guided by the precedent set in~\cite{Cai2020JointCL}. 
In Equation~\ref{equ:PL}, we set the $\gamma$ as $-1$ in all experiments.
Within the scope of Section~\ref{sec:patch_sampling}, which discusses n-nearest entropy-based sampling, the parameter n is determined through validation to hold the values of $0.999$, $0.25$, $0.2$, and $20$, respectively.
For all the baselines, we follow the hyperparameters defined in the original paper, and use Optuna~\cite{optuna_2019} to tune the learning rate.

\subsection{Main Results}

\begin{table}[ht]
\centering
\caption{Comparisons with state-of-the-art semi-supervised learning on LA dataset.}
\label{tab:main_LA}
\resizebox{0.48\textwidth}{!}{
\begin{tabular}{c|c|c|c|c|c|c}
\toprule
\multirow{2}{*}{Method} & \multicolumn{2}{c|}{Scans used}                                & \multicolumn{4}{c}{Metrics}  \\ 
\cline{2-7} 
    & Labeled        & Unlabeled       & DSC$\uparrow$ & Jaccard$\uparrow$ & HD95$\downarrow$ & ASD$\downarrow$ \\ 
\midrule
\multirow{3}{*}{V-Net}  & 5\%  & {0} & 52.55 & 39.60  & 47.05 & 9.87  \\ 
\cline{2-7} 
                        & 10\% & {0} & 82.74  & 71.70 & 13.33 & 3.26 \\ 
\cline{2-7} 
                        & {100\%}                  & {0}  & 91.44  & 84.55 & 5.48  & 1.53 \\ 
\hline
UA-MT                  & \multirow{8}{*}{5\%}  & \multirow{8}{*}{95\%} & 82.26          & 70.98             & 13.71            & 3.82            \\ \cline{4-7}
Double-UA               &                     &                     & 82.73          & 71.73             & 12.53            & 3.80            \\ \cline{4-7} 
SASSNet                 &                     &                     & 81.60          & 69.63             & 16.16            & 3.58            \\ \cline{4-7}
DTC                     &                     &                     & 81.25         & 69.33            & 14.90            & 3.99            \\ \cline{4-7} 
URPC                    &                     &                     & 82.48          & 71.35           & 14.65            & 3.65            \\ \cline{4-7}
MC-Net                  &                     &                     & 83.59          & 72.36             & 14.07            & 2.70            \\ \cline{4-7} 
SS-Net                  &                     &                     & 86.33          & 76.15             & 9.97             & 2.31            \\ \cline{4-7}
ACTION & & &86.60 & 76.20 & 9.70 &2.24 \\ \cline{4-7}
ARCO & & &86.90 & 76.10 & 9.88  &2.73 \\ \cline{4-7}
\textbf{Semi-AGCL}                     &                     &                     & \textbf{90.44} & \textbf{79.05}    & \textbf{7.78}    & \textbf{2.11}   \\
\hline
UA-MT                  & \multirow{8}{*}{10\%} & \multirow{8}{*}{90\%} & 87.79          & 78.39             & 8.68            & 2.12            \\ \cline{4-7} 
Double-UA               &                     &                     & 88.53          & 78.83            & 8.42             & 2.10            \\ \cline{4-7} 
SASSNet                 &                     &                     & 87.54          & 78.05             & 9.84             & 2.59            \\ \cline{4-7} 
DTC                     &                     &                     & 87.51          & 78.17             & 8.23             & 2.36            \\ \cline{4-7} 
URPC                    &                     &                     & 86.92          & 77.03             & 11.13            & 2.28            \\ \cline{4-7}
MC-Net                  &                     &                     & 87.66          & 78.25            & 10.03            & 1.82            \\ \cline{4-7}
SS-Net                  &                     &                     & 88.55          & 79.62             & 7.49             & 1.90            \\ \cline{4-7}
ACTION & & & 88.7 & 78.92 & 8.11 &2.10 \\ \cline{4-7}
ARCO & & &89.1 & 80.71 & 7.78 &2.30 \\ \cline{4-7}
\textbf{Semi-AGCL}                    &                     &                     & \textbf{90.33} & \textbf{82.53}    & \textbf{6.68}    & \textbf{1.78}   \\ \bottomrule
\end{tabular}}
\end{table}

\begin{figure*}[!t]
\centering
\begin{subfigure}[b]{\textwidth}
        \centering
        \includegraphics[width=\textwidth]{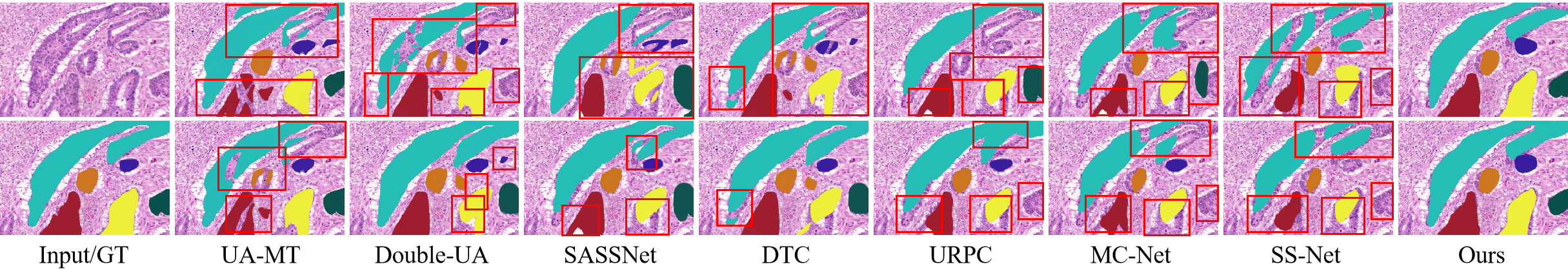}
        \label{fig:viz_crag}
\end{subfigure}

\caption{The visualization of the proposed framework and baselines on the CRAG, LA and  dataset. The first and second rows are the segmentation results with labeled ratios of $5\%$ and $10\%$, respectively. The red boxes indicate that our method outperforms other baselines. GT: Ground Truth.}
\label{fig:visualization}
\end{figure*}

\begin{table}[ht]
\centering
\caption{Comparisons with state-of-the-art semi-supervised learning on the ACDC dataset.}
\label{tab:main_ACDC}
\resizebox{0.48\textwidth}{!}{
\begin{tabular}{c|c|c|c|c|c|c}
\toprule
\multirow{2}{*}{Method} & \multicolumn{2}{c|}{Scans used}                                & \multicolumn{4}{c}{Metrics}                                                                                                           \\ 
\cline{2-7} 
                        & Labeled        & Unlabeled       & DSC$\uparrow$ & Jaccard$\uparrow$ & HD95$\downarrow$ & ASD$\downarrow$ \\ 
\midrule
\multirow{3}{*}{U-Net} & {5\%}                   & 0                   & 47.82          & 37.01          & 31.16         & 12.66         \\ \cline{2-7} 
                       & {10\%}                  & 0                   & 78.22          & 68.05          & 9.33          & 2.70          \\ \cline{2-7} 
                       & {100\%}                  & 0                   & 91.44          & 84.55          & 4.30          & 1.00          \\ 
\hline
UA-MT                 & \multirow{8}{*}{5\%}  & \multirow{8}{*}{95\%} & 46.04          & 35.97          & 20.08         & 7.75          \\ \cline{4-7}
Double-UA              &                     &                     & 56.88          & 45.53          & 22.70         & 6.26          \\ \cline{4-7}
SASSNet                &                     &                     & 57.77          & 46.14          & 20.05         & 6.06          \\ \cline{4-7}
DTC                    &                     &                     & 56.90          & 45.66          & 23.33         & 7.38          \\ \cline{4-7}
URPC                   &                     &                     & 55.58          & 43.66          & 13.66         & 3.78          \\ \cline{4-7}
MC-Net                 &                     &                     & 62.85          & 52.29          & 7.62          & 2.33          \\ \cline{4-7}
SS-Net                 &                     &                     & 65.83          & 55.38          & 6.67          & 2.28          \\ \cline{4-7}
ACTION & & &87.23 &75.34 &2.23 &1.47 \\ \cline{4-7}
ARCO & & &88.51 &76.54 &2.20 &1.40 \\ \cline{4-7}
\textbf{Semi-AGCL}                   &                     &                     & \textbf{88.92} & \textbf{78.84} & \textbf{1.90} & \textbf{0.66} \\ 
\hline
UA-MT                 & \multirow{8}{*}{10\%} & \multirow{8}{*}{90\%} & 81.66          & 70.56          & 6.88          & 2.00          \\ \cline{4-7} 
Double-UA              &                       &                       & 84.48          & 73.97          & 5.52          & 1.90          \\ \cline{4-7} 
SASSNet                &                       &                       & 84.50          & 74.34          & 5.42          & 1.88          \\ \cline{4-7} 
DTC                    &                       &                       & 84.29          & 73.72          & 12.81         & 4.00          \\ \cline{4-7} 
URPC                   &                       &                       & 83.11          & 72.41          & 4.84          & 1.55          \\ \cline{4-7} 
MC-Net                 &                       &                       & 86.47          & 77.13          & 5.50          & 1.83          \\ \cline{4-7} 
SS-Net & & & 86.78          & 77.44          & 6.00          & 1.40          \\ \cline{4-7}
ACTION & & &89.70 &78.86 &4.36 &2.33 \\ \cline{4-7}
ARCO & & &\textbf{92.20} &81.96 &3.44 &2.53 \\ \cline{4-7}
\textbf{Semi-AGCL} & & & 91.98 & \textbf{82.96} & \textbf{3.36} & \textbf{1.16}
\\\bottomrule
\end{tabular}}
\end{table}

\begin{table}[!th]
\caption{Comparisons with state-of-the-art semi-supervised learning on the CARG dataset.}
\label{tab:main_CARG}
\begin{tabular}{c|c|c|c|c|c|c}
\toprule
\multirow{2}{*}{Method} & \multicolumn{2}{c|}{Scans used}                                    & \multicolumn{4}{c}{Metrics}                                                                                                           \\ \cline{2-7} 
                        & \multicolumn{1}{c|}{Labeled}               & Unlabeled             & \multicolumn{1}{c|}{DSC$\uparrow$} & \multicolumn{1}{c|}{Jaccard$\uparrow$} & \multicolumn{1}{c|}{HD95$\downarrow$} & ASD$\downarrow$ \\ 
\midrule
\multirow{3}{*}{U-Net}  & \multicolumn{1}{c|}{{5\%}}                 & 0                     & \multicolumn{1}{c|}{40.77}              & \multicolumn{1}{c|}{33.57}                  & \multicolumn{1}{c|}{30.11}                 &   11.66              \\ \cline{2-7} 
                        & \multicolumn{1}{c|}{{10\%}}                & 0                     & \multicolumn{1}{c|}{75.42}              & \multicolumn{1}{c|}{70.05}                  & \multicolumn{1}{c|}{8.22}                 &  2.82               \\ \cline{2-7} 
                        & \multicolumn{1}{c|}{{100\%}}               & 0                     & \multicolumn{1}{c|}{91.10}         & \multicolumn{1}{c|}{83.28}                  & \multicolumn{1}{c|}{1.19}             & 1.98            \\ \hline
UA-MT                   & \multicolumn{1}{c|}{\multirow{8}{*}{5\%}}  & \multirow{8}{*}{95\%} & \multicolumn{1}{c|}{47.75}              & \multicolumn{1}{c|}{38.81}                  & \multicolumn{1}{c|}{18.44}                 &  6.36               \\ \cline{4-7} 
Double-UA         & \multicolumn{1}{c|}{}      &      & \multicolumn{1}{c|}{50.42}                         & \multicolumn{1}{c|}{44.45}              & \multicolumn{1}{c|}{15.87}                  & {7.05}                            \\ \cline{4-7} 
SASSNet                 & \multicolumn{1}{c|}{}                      &                       & \multicolumn{1}{c|}{48.87}              & \multicolumn{1}{c|}{40.63}                  & \multicolumn{1}{c|}{18.87}                 &  6.77               \\ \cline{4-7} 
DTC                     & \multicolumn{1}{c|}{}                      &                       & \multicolumn{1}{c|}{50.50}              & \multicolumn{1}{c|}{45.60}                  & \multicolumn{1}{c|}{15.92}                 &  6.51               \\ \cline{4-7} 
URPC                    & \multicolumn{1}{c|}{}                      &                       & \multicolumn{1}{c|}{58.85}              & \multicolumn{1}{c|}{48.89}                  & \multicolumn{1}{c|}{13.99}                 &  5.95               \\ \cline{4-7} 
MC-Net                  & \multicolumn{1}{c|}{}                      &                       & \multicolumn{1}{c|}{58.88}              & \multicolumn{1}{c|}{50.50}                  & \multicolumn{1}{c|}{9.50}                 &  5.25               \\ \cline{4-7} 
SS-Net                  & \multicolumn{1}{c|}{}                      &                       & \multicolumn{1}{c|}{58.95}              & \multicolumn{1}{c|}{48.88}                  & \multicolumn{1}{c|}{10.75}                 & 4.95                \\ \cline{4-7} 
ACTION & & &66.43 & 60.13 & 7.40 & 4.10 \\ \cline{4-7}
ARCO & & & 70.63 & 63.33 & 5.20 & 3.33 \\ \cline{4-7}
\textbf{Semi-AGCL}                    & \multicolumn{1}{c|}{}                      &                       & \multicolumn{1}{c|}{\textbf{84.42}}              & \multicolumn{1}{c|}{\textbf{70.49}}                  & \multicolumn{1}{c|}{\textbf{1.48}}                 & \textbf{2.88}                \\ \hline
UA-MT                   & \multicolumn{1}{c|}{\multirow{8}{*}{10\%}} & \multirow{8}{*}{90\%} & \multicolumn{1}{c|}{81.46}         & \multicolumn{1}{c|}{71.42}                  & \multicolumn{1}{c|}{1.48}             & 2.23            \\ \cline{4-7} 
Double-UA               & \multicolumn{1}{c|}{}                      &                       & \multicolumn{1}{c|}{87.01}         & \multicolumn{1}{c|}{77.58}                  & \multicolumn{1}{c|}{1.50}             & 2.63            \\ \cline{4-7} 
SASSNet                 & \multicolumn{1}{c|}{}                      &                       & \multicolumn{1}{c|}{86.43}              & \multicolumn{1}{c|}{76.98}                  & \multicolumn{1}{c|}{1.67}                 &  2.66               \\ \cline{4-7} 
DTC                     &                       &                       & \multicolumn{1}{c|}{84.13}         & \multicolumn{1}{c|}{75.24}                  & \multicolumn{1}{c|}{1.83}             & 2.73            \\ \cline{4-7} 
URPC                    & \multicolumn{1}{c|}{}                      &                       & \multicolumn{1}{c|}{83.36}              & \multicolumn{1}{c|}{71.79}                  & \multicolumn{1}{c|}{1.61}                 &  2.33               \\ \cline{4-7} 
MC-Net                  & \multicolumn{1}{c|}{}                      &                       & \multicolumn{1}{c|}{83.30}              & \multicolumn{1}{c|}{72.11}                  & \multicolumn{1}{c|}{1.61}                 &  2.13               \\ \cline{4-7} 
SS-Net                  & \multicolumn{1}{c|}{}                      &                       & \multicolumn{1}{c|}{83.40}              & \multicolumn{1}{c|}{70.25}                  & \multicolumn{1}{c|}{1.88}                 &  2.58               \\ \cline{4-7} 
ACTION & & &85.56 & 77.33 & 1.55 & 2.05 \\ \cline{4-7}
ARCO & & & 88.81 & 80.90 & 1.33 & 1.88 \\ \cline{4-7}
\textbf{Semi-AGCL}                     & \multicolumn{1}{c|}{}                      &                       & \multicolumn{1}{c|}{\textbf{91.93}}              & \multicolumn{1}{c|}{\textbf{83.37}}                  & \multicolumn{1}{c|}{\textbf{1.08}}                 &  \textbf{1.76}               \\ 
\bottomrule
\end{tabular}

\end{table}

The proposed framework is compared with the state-of-the-art CL- and SemiSL-based segmentation methods at different markers (i.e., $5\%$ and $10\%$), that is, UA-MT~\cite{Yu2019UncertaintyawareSM}, Double-UA~\cite{Wang2020DoubleUncertaintyWM}, SASSNet~\cite{Li2020ShapeawareS3}, DTC~\cite{Luo2020SemisupervisedMI}, URPC~\cite{Luo2020EfficientSG}, MC-Net~\cite{Wu2021SemisupervisedLA}, and SS-Net~\cite{Wu2022ExploringSA}. We also present results with distillation-based semi-supervised learning methods with ACTION~\cite{you2023bootstrapping} and ARCO~\cite{you2023rethinking}. To note that both ACTION and ARCO are built pretrained model and fine-tuning strategies to built the model which involved extensive computational budget. We mannully choose the lowest loss of the pretrained model and then fine-tuning the model with fixed hyperparameters in the orignal papers.

\textbf{LA dataset.}
Results from other competitors are reported in the identical experimental setting in SS-Net~\cite{Wu2022ExploringSA} for fair comparisons. As shown in Table~\ref{tab:main_LA}, our framework achieves the best performance on all four evaluation metrics, significantly outperforming other competitors.
For settings with a $5\%$ labeled ratio, we achieve significant improvements over Dice, Jaccard, HD95, and ASD (i.e., $3.11\%$, $2.90\%$, $2.19$, and $0.20$ over the second one, respectively). This is because of the loss function designed based on the affinity graph, which can directly utilize the inherent structure of the data and encapsulate the geometric and topological relationships between data, helping to enhance intra-class compactness and inter-class separability to improve the framework's ability to learn effective features, thereby improving the segmentation performance of medical images.
We show the segmentation results in the supplementary material.

\begin{table*}[ht]
\centering
\caption{Comparisons with different setting of affinity graph loss on the ACDC dataset. GK+L* = $\sum\limits^N_{i=1} \exp\left(-\frac{L*}{2\sigma^2}\right)$ in Equ.~\ref{equ:PL}.}
\label{tab:AGG_loss}
\resizebox{0.9\textwidth}{!}{
\begin{tabular}{*{13}{c|}c}
\toprule
\multicolumn{3}{c|}{$\mathcal{L}^{PL}_{AGG}$}                                                                   & \multicolumn{3}{c|}{$\mathcal{L}^{RW}_{AGG}$}                                                & \multicolumn{4}{c|}{Label=5\%}                                                                                 & \multicolumn{4}{c}{Label=10\%}                                                                                \\ \hline
\multicolumn{1}{c|}{GK+L2} & \multicolumn{1}{c|}{GK+L1} & \multicolumn{1}{c|}{$\lambda||\mathbf{A}||*$} & \multicolumn{1}{c|}{$A_{ii}$} & \multicolumn{1}{c|}{1:1} & \multicolumn{1}{c|}{random} & DSC$\uparrow$ & Jacard$\uparrow$ & \multicolumn{1}{c|}{HD95$\downarrow$} & ASD$\downarrow$ & DSC$\uparrow$ & \multicolumn{1}{c|}{Jacard$\uparrow$} & \multicolumn{1}{c|}{HD95$\downarrow$} & \multicolumn{1}{c}{ASD$\downarrow$} \\ \midrule
-                           & -                           & -                           & -          & -          & -          & 68.82          & 68.77          & 8.67          & 2.40          & 86.78          & 77.63          & 6.68          & 2.00          \\ \hline
\multirow{8}{*}{\checkmark} & \multirow{8}{*}{}           & \multirow{4}{*}{\checkmark} &            &            &            & 66.43          & 68.78          & 8.58          & 2.23          & 87.13          & 77.53          & 6.21          & 1.90          \\ \cline{4-14} 
                            &                             &                             & \checkmark &            &            & \textbf{88.92} & \textbf{78.84} & \textbf{1.90} & \textbf{0.66} & 89.98          & \textbf{80.96} & \textbf{3.66} & \textbf{1.16} \\ \cline{4-14} 
                            &                             &                             &            & \checkmark &            & 86.06          & 75.83          & 7.88          & 1.93          & 87.93          & 78.83          & 8.44          & 2.00          \\ \cline{4-14} 
                            &                             &                             &            &            & \checkmark & 85.56          & 73.26          & 8.53          & 2.33          & 88.10          & 79.82          & 7.78          & 1.83          \\ \cline{3-14} 
                            &                             & \multirow{4}{*}{}           &            &            &            & 67.78          & 68.88          & 8.54          & 2.89          & 87.78          & 77.56          & 6.70          & 2.10          \\ \cline{4-14} 
                            &                             &                             & \checkmark &            &            & 71.71          & 72.53          & 7.74          & 2.11          & 87.78          & 77.58          & 6.71          & 2.15          \\ \cline{4-14} 
                            &                             &                             &            & \checkmark &            & 70.71          & 71.33          & 8.21          & 2.10          & 87.72          & 77.50          & 5.93          & 1.98          \\ \cline{4-14} 
                            &                             &                             &            &            & \checkmark & 69.83          & 70.88          & 8.11          & 2.00          & 87.78          & 77.47          & 6.32          & 1.98          \\ \hline
\multirow{8}{*}{}           & \multirow{8}{*}{\checkmark} & \multirow{4}{*}{\checkmark} &            &            &            & 66.52          & 69.00          & 7.93          & 2.07          & 85.55          & 74.32          & 7.01          & 2.11          \\ \cline{4-14} 
                            &                             &                             & \checkmark &            &            & 87.93          & 78.63          & 2.33          & 1.39          & \textbf{90.12} & \textbf{80.96} & 3.68          & 1.19          \\ \cline{4-14} 
                            &                             &                             &            & \checkmark &            & 87.78          & 75.53          & 3.53          & 1.66          & 86.59          & 74.44          & 7.43          & 2.31          \\ \cline{4-14} 
                            &                             &                             &            &            & \checkmark & 79.35          & 71.33          & 7.01          & 1.99          & 85.58          & 75.00          & 6.66          & 2.13          \\ \cline{3-14} 
                            &                             & \multirow{4}{*}{}           &            &            &            & 66.52          & 69.10          & 7.53          & 2.01          & 85.75          & 74.42          & 6.83          & 1.89          \\ \cline{4-14} 
                            &                             &                             & \checkmark &            &            & 83.10          & 73.11          & 5.56          & 1.66          & 85.72          & 74.58          & 5.38          & 1.66          \\ \cline{4-14} 
                            &                             &                             &            & \checkmark &            & 82.33          & 70.31          & 6.87          & 1.77          & 85.22          & 74.46          & 6.82          & 1.89          \\ \cline{4-14} 
                            &                             &                             &            &            & \checkmark & 82.33          & 70.52          & 7.05          & 1.86          & 85.40          & 74.23          & 6.50          & 1.84          \\ 
\bottomrule
\end{tabular}}
\end{table*}

\textbf{ACDC dataset.}
Following SS-Net~\cite{Wu2022ExploringSA}, we use 2D U-Net as the backbone, set the input patch size as $256 \times 256$ and the size of the zero-value region of mask $\mathcal{M}$ as $170 \times 170$. The batch size, pre-training iterations, and the self-training training iterations are set as $24$, $10$k and $30$k, respectively.
Table~\ref{tab:main_ACDC} shows the averaged performance of four-class segmentation results on ACDC dataset with $5\%$ and $10\%$ labeled ratios.
It can be seen that our method is clearly optimal, e.g., with $5\%$ labeled ratio, we obtain a huge performance improvement of up to $23.09\%$ in DSC.
The HD95 and ASD in the $10\%$ labeled ratio have decreased compared to that of $5\%$, which may be because the loss function based on the affinity graph requires an appropriate increase in training iterations for the complex details of the edges. However, overall, our method significantly outperforms the competition on all metrics for all labeled ratios.
We show the segmentation results in the supplementary material.

\textbf{CARG dataset.}
We follow~\cite{shi2022semi} to split the data into $80-10-10$ training, test, and validation ratios.
Table~\ref{tab:main_CARG} shows that our method achieves great improvement in the CARG dataset, and even at $10\%$ labeled ratio, our method performs better segmentation than U-Net with $100\%$ labeled ratio.
This is mainly because
(i) 2D slices can generate more combinations than combinations of 3D data. Therefore, knowledge from labeled data can be more fully transferred to unlabeled data, especially when the amount of labeled data is extremely small. This may be the reason why such a significant improvement is achieved when the labeled ratio is $5\%$ compared to $10\%$ (the ACDC dataset also has this advantage).
And (ii) the CARG dataset has WS histopathological images, which contain additional texture features that can enrich the edge information for the affinity graph. This may be the reason why the improvement performance in the CARG dataset is more obvious than that of other datasets.
Besides, as can be seen from Figure~\ref{fig:visualization}, our method can accurately segment all objects, and the segmentation details are closer to ground truths than other baselines (see the red boxes).

\begin{table}[!ht]
\caption{Evaluation of diverse similarity metrics in patch sampling on the ACDC dataset.}
\label{tab:patch}
\resizebox{0.48\textwidth}{!}{
\begin{tabular}{c|c|c|c|c|c|c|c|c}
\toprule
\multirow{2}{*}{Similarity} & \multicolumn{4}{c|}{Label = 5\%}& \multicolumn{4}{c}{Label = 10\%}\\ \cline{2-9} 
                            & \multicolumn{1}{c|}{DSC$\uparrow$}  & \multicolumn{1}{c|}{Jacard$\uparrow$} & \multicolumn{1}{c|}{HD95$\downarrow$} & ASD$\downarrow$ & \multicolumn{1}{c|}{DSC$\uparrow$}  & \multicolumn{1}{c|}{Jacard$\uparrow$} & \multicolumn{1}{c|}{HD95$\downarrow$} & ASD$\downarrow$ \\ \midrule
Cosine           & 74.32          & 66.13            & 10.86            & 3.87            & 81.07          & 70.35            & 9.23             & 6.77            \\ \hline
Class Confidence & 78.85          & 69.20            & 10.02            & 3.04            & 85.51          & 74.74            & 5.09             & 3.11            \\ \hline
Entropy (Ours)   & \textbf{88.92} & \textbf{78.84}   & \textbf{1.90}    & \textbf{0.66}   & \textbf{89.98} & \textbf{80.96}   & \textbf{3.66}    & \textbf{0.86}  \\
\bottomrule

\end{tabular}
}
\end{table}

\begin{table}[h]
\centering
\caption{
Quantitative comparison of computational time between our methods and other semi-supervised learning methods on Left Atrium MRI dataset. We also present the Semi-AGCL without patch-wise class centric sampling (see Semi-AGCL w/cls conf). The Params is refer to the number of trainable parameters using the same backbone.  \label{tab:comp}}
\resizebox{0.48\textwidth}{!}{
\begin{tabular}{c|c|c|c|c}
\toprule
\multirow{2}{*}{Method} & \multicolumn{2}{c|}{Scaned Used} & \multicolumn{2}{c}{Computational Cost}         \\ \cline{2-5} 
      & \multicolumn{1}{c|}{Labeled}              & Unlabeled             & \multicolumn{1}{c|}{Params (M)} & Training time (mins) \\ \hline
\multirow{2}{*}{VNet}   & \multicolumn{1}{c|}{5\%}    & 0  & \multicolumn{1}{c|}{9.44}  & 36.5 \\ \cline{2-5} 
                        & \multicolumn{1}{c|}{100\%}  & 0  & \multicolumn{1}{c|}{9.44}  & 37.8 \\ \hline
UA-MT & \multicolumn{1}{c|}{\multirow{6}{*}{5\%}} & \multirow{6}{*}{95\%} & \multicolumn{1}{c|}{9.44}       & 67.5             \\ \cline{1-1} \cline{4-5} 
SASSNet                 &        &    & 20.46 & 73.6 \\ \cline{1-1} \cline{4-5} 
DTC                     &        &    & 9.44  & 47.1 \\ \cline{1-1} \cline{4-5} 
MC-Net                  &        &    & 15.25 & 88.9 \\ \cline{1-1} \cline{4-5} 
SS-Net                  &        &    & 9.44  & 70.8 \\ \cline{1-1} \cline{4-5}
ACTION                  &        &    & 10.14  & 471.9   \\ \cline{1-1} \cline{4-5}
ARCO                  &        &    &10.14   & 421.1  \\ \cline{1-1} \cline{4-5}
Semi-AGCL(Ours)         &        &    & \multicolumn{1}{c|}{9.44}  & 48.6 \\ \cline{1-1} \cline{4-5}
Semi-AGCL w/cls conf         &        &    & \multicolumn{1}{c|}{9.44}  & 46.8 \\

\bottomrule
\end{tabular}
}
\end{table}

\textbf{Computational Efficiency}
Since most of semi-supervised learning methods are constructed based on complicated pipeline setup, we present the quantitative comparison of network’s parameters and training time are listed in Table~\ref{tab:comp} on LA dataset. For all the semi-supervisd learning pipeline, our proposed Semi-AGCL achieved the second best training time over all existing semi-supervised learning methods. However, the performance of Semi-AGCL is much more better then DTC in LA datasets.
Although our proposed method involves extensive matrix manipulation, it is highly parallelizable, providing an advantage in computation time. Furthermore, the patch-wise class-centric sampling method does not require parameter tuning. Despite the complexity of this computation, it did not substantially increase our computation time. As shown in Table~\ref{tab:comp}, the training time difference between Semi-AGCL with patch-wise class-centric sampling and Semi-AGCL with class confidence (cls conf) sampling is only 1.8 minutes.

\subsection{Ablation Studies}

\textbf{Effectiveness of each module.}
In Table~\ref{tab:AGG_loss}, under different labeling ratios, we compare different loss functions between pseudo-labels and loss functions based on different positive and negative sample selection methods, proving the effectiveness and optimality of the proposed loss functions (i.e., $\mathcal{L}^{PL}_{AGG}$ and $\mathcal{L}^{RW}_{AGG}$).
First, for $\mathcal{L}^{PL}_{AGG}$, we find that the experimental results of $GK+L*$ used to calculate $A_{ij}$ are not much different, but adding $\lambda||A||*$ used to calculate $A_{ii}$, the segmentation results have improved (the $random$ of $\mathcal{L}^{RW}_{AGG}$ does not meet this conclusion, which may be because the randomness of the selection of samples is too high).
Then, for $\mathcal{L}^{RW}_{AGG}$, the designed sampling method ($A_{ii}$) is better than the other two methods in most cases, but it incorporating $\lambda||A||*$ will significantly improve all metrics. These phenomena prove the optimality and complementarity of our loss function and sampling method based on the affinity graph.
Finally, we find that the improvement under the labeling ratio of $5\%$ is more obvious than that of $10\%$, further proving the effectiveness of our method in extreme data.

\textbf{Effectiveness of the patch-wise class-centric sampling.}
The comparison is conducted among three patch sampling methods: Cosine Similarity, Class Confidence, and our proposed entropy-based technique.
Cosine similarity 
is a prevalent metric for gauging similarity between two vectorized patches. 
The class confidence for a given patch $P^k_{i, j}$ requires computing the average patch confidence and subsequently classifying patches with analogous confidence values as positive and the rest as negative. 
We observe from Table~\ref{tab:patch} that the cosine similarity-centric patch-sampling from $X_i^{'k}$ is not satisfactory. Class confidence only achieves limited improvement relative to cosine similarity.
This is mainly because the classification of positive and negative samples is not perpetually exclusive, and the misclassification rate of the above methods may increase, resulting in suboptimal performance.
Our method achieves huge improvements,
this is because our method advocates using entropy in $X_i^{'k}$ to sample positive and negative patches according to the class confidence map, and considers it as a more efficient measure for disparity mapping amidst patches.


\section{Conclusion}
To alleviate the problem that methods combining semi-supervised learning and contrastive learning rely on pretext tasks and insufficient supervision signals, we propose an affinity-graph-guided semi-supervised contrastive learning framework (Semi-AGCL) to achieve medical image segmentation without pretext under extremely few annotations.
Semi-AGCL first designs an average-patch-entropy-driven inter-patch sampling method, which can provide a powerful initial feature space without pretext tacks; and then it designs a new affinity-graph-based loss function between the student and teacher networks to improve the model's generalization ability.
Evaluation on three medical segmentation datasets spanning multiple domains, our framework outperforms SOTA methods with minimal annotations, confirming its effectiveness and generalizability.

\bibliographystyle{ieeetr}
\bibliography{refer}
\end{document}